%% file: curvefp_arxiv.tex
\documentclass[11pt]{article}

\input{math_commands.tex}

\usepackage{booktabs}
\usepackage{graphicx}
\usepackage{hyperref}
\usepackage{microtype}
\usepackage{multirow}
\usepackage{natbib}
\usepackage{placeins}
\usepackage{times}
\usepackage{url}
\usepackage{xcolor}
\usepackage{xspace}

\definecolor{curveblue}{HTML}{2463A6}
\definecolor{curvegreen}{HTML}{16856C}
\hypersetup{
    colorlinks=true,
    linkcolor=curveblue,
    citecolor=curvegreen,
    urlcolor=curveblue,
    pdftitle={CurveFP: Rational-Radix Logarithmic Datatypes with Closed Products for Language Models},
    pdfauthor={Ye Qiao}
}

\newcommand{\curvefp}{\textsc{CurveFP}\xspace}
\newcommand{\curvefpeight}{\textsc{CurveFP8}\xspace}
\newcommand{\curvefpseven}{\textsc{CurveFP7}\xspace}

\title{\curvefp: Co-Designing Numerical Representation \\ and Product Arithmetic for Language Models}
\author{Ye Qiao\\
Department of Electrical Engineering and Computer Science\\
University of California, Irvine\\
Irvine, CA, USA}
\date{}

\begin{document}

\sloppy

\maketitle

\begin{abstract}
Low-precision formats usually optimize scalar fidelity while inheriting
conventional product arithmetic. We introduce \curvefp, a block-scaled family
that distributes magnitudes across interleaved logarithmic curves. Uniform
curve indices make every nonzero product an exact sign and integer-index update,
while a rational radix exposes the finite phase schedule required for
accumulation. We instantiate the algebra as \curvefpeight E4C3/E5C2 for
training and \curvefpseven E3C3 for compact inference. On four 7B--9B models,
\curvefpseven beats tensorwise FP8 perplexity with one fewer element bit and
stays within 1.32\% of native quality. \curvefpeight lowers error in all 36
paired training-GEMM comparisons. Across three matched 3B-token pretraining
triplets, it reaches mean BF16-inference perplexity 22.5366 versus 22.5407 for
FP8 and has a lower format penalty in every seed. Downstream evaluation shows
transfer parity and a consistent WikiText-103 gain. In a preliminary
$4\times4$ Nangate45 spatial tile, \curvefpeight uses one fewer product
register and 4.6\% less area than timing-closing FP8 at 500 MHz. These results
support \curvefp as a numerical and arithmetic co-design, while leaving
system-level efficiency to future study.
\end{abstract}

\input{sections/introduction}
\input{sections/related_work}
\input{sections/method}
\input{sections/experiments}
\FloatBarrier
\input{sections/discussion}
\input{sections/conclusion}

\section*{Reproducibility Statement}
Section~\ref{sec:method} defines the formats, quantization rule, product mapping,
and phase count. Section~\ref{sec:experiments} reports the model, data,
optimizer, task, and precision contracts used in each principal experiment.
The supplementary code fixes the seed and data contracts, implements resumable
checkpointing and atomic artifact generation, validates every matrix cell, and
regenerates result summaries and figures from produced artifacts.
Section~\ref{sec:deployment} and Appendix~\ref{app:hardware} report the
preliminary spatial-tile realization, verification, and open tool flow.

\section*{AI Use Statement}
We used generative AI tools to assist with experiment scripts, derivation
checking, figure preparation, literature review, and language editing.

\bibliography{iclr2027_conference}
\bibliographystyle{iclr2027_conference}

\clearpage
\appendix
\input{sections/appendix}

\end{document}

%% file: math_commands.tex
\usepackage{amsmath,amsfonts,bm}

\def\eqref#1{equation~\ref{#1}}
\def\1{\bm{1}}

\DeclareMathAlphabet{\mathsfit}{\encodingdefault}{\sfdefault}{m}{sl}
\SetMathAlphabet{\mathsfit}{bold}{\encodingdefault}{\sfdefault}{bx}{n}

%% file: sections/introduction.tex
\section{Introduction}
\label{sec:introduction}

Low-precision arithmetic has become a primary lever for scaling language-model
training and inference, yet prevailing datatypes optimize how individual values
are approximated while largely accepting the cost of their induced products.
FP8 established that role-specific exponent and mantissa layouts can match
16-bit training quality~\citep{micikevicius2022fp8}; microscaling formats then
amortized range through per-block scales~\citep{rouhani2023microscaling}.
Post-training methods attack the same numerical bottleneck from the data side,
moving activation outliers into weights~\citep{xiao2023smoothquant} or rotating
them away~\citep{ashkboos2024quarot,liu2025spinquant}. These advances make
quantization substantially more accurate, but their element products still
require a conventional integer or floating-point multiplier.

Product structure is an underused source of efficiency in low-precision
design. Learned product/subspace codebooks offer flexible compression
points~\citep{wang2026apexq,qiao2026fasq}, but do not by themselves impose a
scalar closed-product algebra. Uniform integers simplify
accumulation but depend increasingly on fine-grained scales as activation range
widens. Logarithmic representations make multiplication additive, but coarse
powers of two sacrifice local resolution and still require an accumulation
strategy~\citep{miyashita2016logarithmic}. Additive powers-of-two codes recover
more levels through several shift-add terms~\citep{li2020apot}. These trade-offs
motivate a stronger objective than scalar error alone: jointly design the level
geometry and product algebra so that lower precision improves both
representation density and arithmetic regularity.

\begin{figure*}[t]
    \centering
    \includegraphics[width=0.98\textwidth]{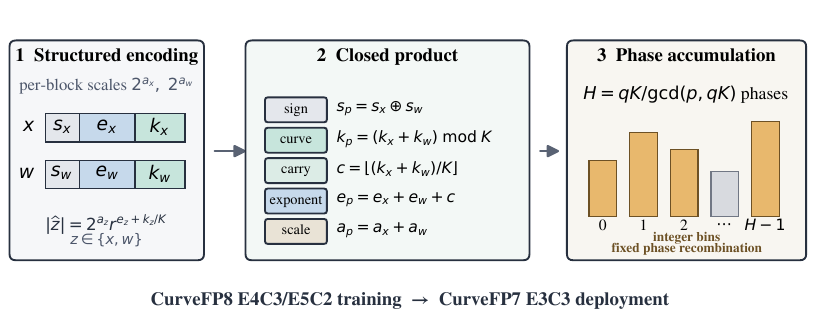}
    \caption{\textbf{\curvefp co-designs representation and arithmetic.}
    Power-of-two scales provide block range; sign, exponent, and curve fields
    encode each element. Uniform curve indices make products exactly closed,
    and the rational radix determines the fixed-weight accumulation phases.}
    \label{fig:curvefp-overview}
\end{figure*}

We realize this objective with \curvefp, a block-scaled closed-product codebook
family. As illustrated in Figure~\ref{fig:curvefp-overview}, \curvefp spreads
quantized magnitudes across $K$ interleaved logarithmic curves. Each nonzero
element stores a sign, a two's-complement exponent, and a curve index under a
shared power-of-two scale. Uniform curve spacing makes multiplication exact in
the code domain: sign XOR, curve-index addition, and exponent addition with a
carry. A reduced rational radix $r=2^{p/q}$ tunes dynamic range against local
precision, while the derived phase count $qK/\gcd(p,qK)$ exposes the number of
fractional exponent classes that accumulation must combine. Thus every
range--resolution choice has an explicit accumulation contract, and every
representable product lands on a known phase without lookup, projection, or a
general product-forming multiplier.

One algebra supports two practical operating points. \curvefpeight uses E4C3
for forward weights and activations and E5C2 for backward gradients, directly
mirroring the range allocation of FP8 E4M3/E5M2 while replacing mantissa
products with closed curve products. \curvefpseven uses E3C3 to reduce the
element width by 12.5\% for deployment while retaining eight product phases.
Block scaling supplies the local range that seven-bit activations need; compact
binary scale exponents preserve the shift-and-count accumulation structure.
A single numeric family therefore spans training and inference with the same
closure rule.

The evidence spans model quality and arithmetic feasibility. Block-scaled
\curvefpseven beats tensorwise FP8 perplexity on four 7B--9B models with one
fewer element bit and remains within 1.32\% of native quality. \curvefpeight
lowers error in all 36 paired forward and backward GEMM windows. Across three
matched 3B-token triplets, its mean BF16-inference perplexity is 22.5366 versus
22.5407 for FP8, with a lower format-induced penalty in every seed; the seed
variation supports parity rather than statistical superiority. A preliminary
matched $4\times4$ spatial tile closes at 500 MHz with one fewer product
register and 4.4\% less area than FP8. Restoring that stage leaves
\curvefpeight 3.9\% smaller and provides additional setup slack, exposing a
useful area--pipeline trade-off.

Our contributions are:
\begin{itemize}
    \item We introduce a block-scaled codebook family distributed across
    rational-radix logarithmic curves, make its curve indices exactly closed
    under multiplication, and derive the finite phase count connecting
    representation choice to accumulation.
    \item We instantiate one algebra as \curvefpeight for FP8-class training
    and \curvefpseven for seven-bit deployment, with compact power-of-two scale
    metadata and no general multiplier in product formation.
    \item We provide end-to-end evidence spanning four-model inference,
    rational-radix and scale ablations, all three training GEMMs, three matched
    3B-token runs, a complete downstream transfer study, and a preliminary
    routed spatial-tile comparison with FP8.
\end{itemize}

%% file: sections/related_work.tex
\section{Related Work}
\label{sec:related-work}

\paragraph{Low-precision datatypes.}
FP8 combines E4M3 forward operands with E5M2 gradients to retain training
range and has become the principal eight-bit floating-point
baseline~\citep{micikevicius2022fp8}. Microscaling extends this approach by
sharing scale factors across blocks of narrow floating-point or integer
elements~\citep{rouhani2023microscaling}; hybrid block floating point likewise
shares exponents during DNN training but retains linear
significands~\citep{drumond2018hbfp}. At the accelerator level, TeLLMe couples
ternary weights with table-lookup matrix multiplication for end-to-end LLM
prefill and decode on edge FPGAs~\citep{qiao2026tellme}.

\paragraph{Logarithmic neural-network arithmetic.}
Classical LNS encodes scalar magnitudes in the log domain, making
multiplication additive; neural-network work has applied it to quantized
CNNs/LSTMs, accelerator arithmetic, and low-precision training with
multiplicative updates~\citep{miyashita2016logarithmic,kouretas2018lns,
christ2022lns,zhao2022lnsmadam}. Flattening CurveFP's exponent and curve fields
into $n=Ke+k$ gives a finite block-scaled LNS lattice, which explains the
shared additive product algebra. CurveFP uses that algebra through a different
datatype interface: an explicit multi-curve codebook, compact scale contracts,
the $H=qK/\gcd(p,qK)$ phase law, and role-specific LLM formats trained with
conventional AdamW. APoT instead improves resolution through multiple
shift-add terms~\citep{li2020apot}.

\paragraph{LLM quantization and structured codebooks.}
Modern LLM post-training quantization often reshapes tensors before applying a
conventional datatype. SmoothQuant migrates activation outliers into
weights~\citep{xiao2023smoothquant}; QuaRot and SpinQuant use fixed or learned
rotations to suppress outliers~\citep{ashkboos2024quarot,liu2025spinquant}.
Learned representations offer a complementary route: LO-BCQ clusters blocks
and assigns each cluster an optimized scalar codebook~\citep{elangovan2025lobcq},
while NestQuant uses self-similar nested lattices for low-error matrix
products~\citep{savkin2025nestquant}. These methods target aggressive W4A4
inference and can exceed simple scalar formats in accuracy. Product/subspace
quantization follows another path: APEX-Q permits arbitrary subvector
dimensions, while FASQ offers calibration-free flexible compression points and
custom GPU kernels~\citep{wang2026apexq,qiao2026fasq}. Our objective is
different: \curvefp parameterizes its levels with a rational radix and uniform
curve indices, obtaining a compact closed-product algebra shared by inference
and training. This structure reaches the FP8 quality regime without
calibration-time transformations.

%% file: sections/method.tex
\section{CurveFP}
\label{sec:method}

\paragraph{Overview.}
\curvefp starts from a closed-product codebook view: distribute values across
structured curves so representation quality and product composition improve
together. We first define the rational-radix element code, then derive closed
product formation and phase-structured accumulation, and finally instantiate
role-specific formats for training and compact inference.

\subsection{A rational-radix datatype family}
\label{sec:method:representation}

\curvefp represents a tensor block with one power-of-two scale and represents
each element by a sign, an exponent, and a curve index. Let $E$ and $C$ denote
the exponent and curve-index widths, let $K=2^C$, and choose a reduced rational
radix $r=2^{p/q}$. A nonzero quantized value in block $b$ is
\begin{equation}
    \widehat{x}
    = (-1)^s 2^{a_b} r^{e+k/K},
    \qquad
    e\in\{-2^{E-1},\ldots,2^{E-1}-1\},
    \quad k\in\{0,\ldots,K-1\},
    \label{eq:curvefp-value}
\end{equation}
where $a_b$ is a signed shared-scale exponent. The element therefore occupies
$1+E+C$ bits. We reserve magnitude index zero as a sentinel and omit the
smallest nonzero exponent/curve combination, leaving
$2^{E+C}-1$ positive magnitudes; values outside the scaled range map to an
endpoint. Unlike a learned codebook, the levels require no table of arbitrary
constants: adjacent curve indices differ by the fixed ratio $r^{1/K}$, and
adjacent exponents differ by $r$.

Given $a_b$, quantization preserves the source sign and selects the magnitude
whose reconstructed value is nearest in absolute error, with saturation at the
endpoints. Static operands choose a representable power-of-two scale by
reconstruction-MSE search; dynamic training operands use ceil-absmax scaling.
We write element-wise tensor quantization as $Q(\bm{X})$, so
$Q(\bm{X})_{ij}=\widehat{x}_{ij}$.

The rational radix parameterizes the spacing of this uniform logarithmic
lattice. Setting $p=q=1$ gives levels $2^{e+k/K}$. Choosing $p/q<1$ contracts
the range and increases local resolution without changing the element width,
whereas $p/q>1$ expands the range. \curvefp associates every spacing with the
phase count derived below, exposing a direct quality--complexity control.

\subsection{Closed products and phase-structured accumulation}
\label{sec:method:products}

Uniform curve indices make the product codebook exactly closed. For two
nonzero operands with curve indices $i$ and $j$, define
\begin{equation}
    \ell=(i+j)\bmod K,
    \qquad c=\left\lfloor\frac{i+j}{K}\right\rfloor.
    \label{eq:curve-index-product}
\end{equation}
Their product has sign $s_x\oplus s_w$, shared-scale exponent $a_x+a_w$,
element exponent $e_x+e_w+c$, and curve index $\ell$. Product formation thus
uses an XOR and small integer additions, with no general
variable-by-variable multiplier and no nearest-code projection. A zero sentinel
short-circuits the product to zero without applying the index update. This
closed product path feeds a fixed phase reduction with binary scale transport;
Section~\ref{sec:deployment} summarizes its deployment implications.

Closure holds on widened product coordinates. Requantizing to a finite $E$-bit
destination may saturate; no codebook projection is needed before that explicit
output requantization.

Rational radices preserve this closure while changing the number of binary
phases needed by a dot product. From Equation~\ref{eq:curvefp-value}, every
unscaled magnitude is
\begin{equation}
    r^{e+k/K}=2^{p(eK+k)/(qK)}.
\end{equation}
Let $N=qK$. As the integer $n=eK+k$ advances, the fractional exponent is the
residue $pn\bmod N$ divided by $N$. The additive subgroup generated by $p$ in
$\mathbb{Z}_N$ has order
\begin{equation}
    H=\frac{qK}{\gcd(p,qK)}
    \label{eq:phase-count}
\end{equation}
and therefore defines $H$ phase classes on the underlying lattice. A finite
format may leave some classes empty, but every dot product can be scheduled over
these $H$ phases and written as
\begin{equation}
    \widehat{\bm{x}}^{\mathsf T}\widehat{\bm{w}}
    =\sum_{h=0}^{H-1}2^{h/H}\sum_t n_{h,t}2^t,
    \label{eq:binned-accumulation}
\end{equation}
where each $n_{h,t}$ is a signed integer count. Products are routed exactly to
integer counts indexed by binary shift; only the $H$ fixed phase weights remain
after reduction.

\paragraph{Constructive arithmetic realization.}
Let $g=\gcd(p,qK)$, $n_x=e_xK+k_x$, $n_w=e_wK+k_w$, and
$u=p(n_x+n_w)$. Each nonzero lane increments $A_{h,t}$ by
$\delta=(-1)^{s_x\oplus s_w}$ at $h=(u\bmod qK)/g$ and
$t=a_x+a_w+\lfloor u/(qK)\rfloor$; zero suppresses the update. For binary
radix, $h=(k_x+k_w)\bmod K$ and
$t=a_x+a_w+e_x+e_w+\lfloor(k_x+k_w)/K\rfloor$. Product formation therefore
uses only zero detection, sign XOR, one $(C+1)$-bit curve-index addition, and
signed exponent addition. A length-$D$ dot needs
$1+\lceil\log_2(D+1)\rceil$-bit signed counters. Equal addresses can be merged
with signed popcounts in a histogram design, or shifted directly into one wide
accumulator per phase. Both layouts produce the same phase sums before the
fixed-coefficient reduction and final requantization. Appendix~\ref{app:algorithm:accumulation}
gives the subgroup proof and full state bound.

\subsection{Shared scales, operating points, and training protocol}
\label{sec:method:formats}

Shared scales let the element code focus on within-block variation while
amortizing range metadata. For a block of $G$ values and an $S$-bit scale
exponent, the effective storage rate is
\begin{equation}
    R = 1+E+C+\frac{S}{G}\quad\text{bits per value}.
    \label{eq:effective-rate}
\end{equation}
Because $2^{a_b}$ is itself binary, it shifts the exponent-bin index without
introducing another product phase. We evaluate tensorwise, rowwise, and
fixed-size block scaling, but use block scaling as the compact deployment
contract because activations require local range adaptation.

Table~\ref{tab:curvefp-formats} gives the principal operating points. The
eight-bit training pair follows the familiar split between a precision-oriented
forward format and a range-oriented gradient format. \curvefpeight E4C3
quantizes weights and activations in the forward linear layers, while E5C2
quantizes the output gradients used by both activation- and weight-gradient
matrix products. Moving one bit from $C$ to $E$ preserves 127 positive levels,
expands gradient range, and reduces the phase count from eight to four; this
mirrors the precision-oriented E4M3/range-oriented E5M2 FP8 split. Parameters
and optimizer states remain full precision, and
nonlinear operations execute in bfloat16. \curvefpseven E3C3 keeps eight curve
phases in a seven-bit element for compact post-training deployment.

\begin{table}[t]
    \caption{Principal \curvefp formats. Positive levels exclude zero; $H$ is
    the phase count from Equation~\ref{eq:phase-count}.}
    \label{tab:curvefp-formats}
    \centering
    \small
    \begin{tabular}{lcccccl}
        \toprule
        Format & Bits & $E$ & $C$ & Positive & $H$ & Role \\
        \midrule
        \curvefpseven E3C3 & 7 & 3 & 3 & 63  & 8 & Compact inference \\
        \curvefpeight E4C3 & 8 & 4 & 3 & 127 & 8 & Forward training \\
        \curvefpeight E5C2 & 8 & 5 & 2 & 127 & 4 & Backward gradients \\
        \bottomrule
    \end{tabular}
\end{table}

Training quantizes the operands of every non-head linear layer. The forward
pass evaluates $Q_A(\bm{X})Q_W(\bm{W})^{\mathsf T}$. The backward pass reuses
the quantized forward operands and quantizes the output gradient
$Q_G(\nabla\bm{Y})$, yielding $Q_G(\nabla\bm{Y})Q_W(\bm{W})$ for
$\nabla\bm{X}$ and $Q_G(\nabla\bm{Y})^{\mathsf T}Q_A(\bm{X})$ for
$\nabla\bm{W}$. E4C3 supplies forward weights and activations, while E5C2
supplies output gradients to both backward GEMMs. Parameters, optimizer states,
nonlinear operations, and the tied output head remain at higher precision.

%% file: sections/experiments.tex
\section{Experiments}
\label{sec:experiments}

\subsection{Experimental design}
\label{sec:experiments:design}

We evaluate \curvefp across post-training inference, all three training GEMMs,
matched from-scratch pretraining, and downstream format transfer. The baselines
are BF16/FP16, standard FP8 E4M3/E5M2~\citep{micikevicius2022fp8}, microscaling
formats, and signed integers with practical real-valued scales. Unless stated
otherwise, we quantize every linear layer except the tied vocabulary head and
keep normalization and nonlinear operations in BF16/FP16. Every comparison
shares data, tokenizer, model weights or initialization, and evaluation code;
reported values come from saved artifacts. Appendix~\ref{app:experiments}
provides the complete protocols.

\subsection{CurveFP7 provides an INT8-class inference alternative at seven bits}
\label{sec:experiments:inference}

We first quantize weights and activations of Llama-3-8B, Qwen3-8B,
Qwen3.5-9B, and Falcon-H1-7B~\citep{grattafiori2024llama3,yang2025qwen3,
qwenteam2026qwen35,zuo2025falconh1}. Every run scores the complete WikiText-2
test split with the same 2,048-token rolling-loglikelihood protocol. Within a
model, methods share weights, output-head exclusion, and K/V precision; Llama
and Qwen3 keep dense K/V at common tensorwise FP8 to isolate weight/activation
arithmetic. We compare tensorwise and rowwise FP8 E4M3 with \curvefpseven E3C3
under compact binary scales. Including scales, G64 costs 7.062 weight and 7.078
activation bits per value; G512 reduces these rates to 7.008 and 7.010.

\begin{figure*}[t]
    \centering
    \includegraphics[width=0.96\textwidth]{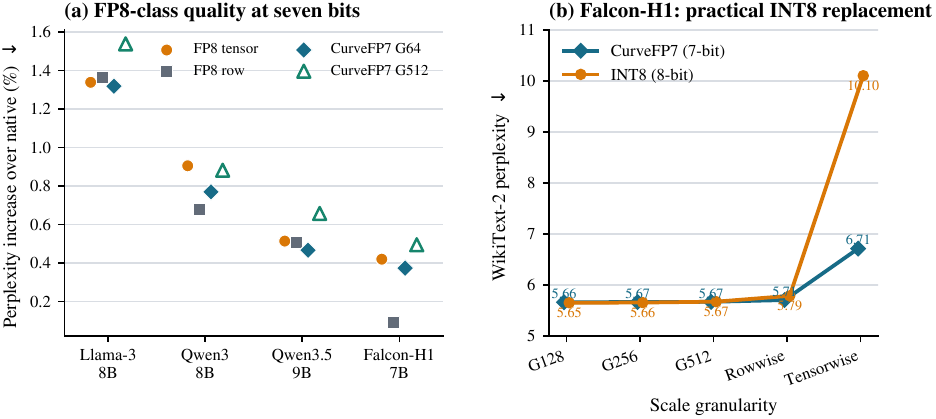}
    \caption{\textbf{Seven-bit \curvefpseven reaches the FP8 quality regime.}
    Left: perplexity increase over native on four models. Right: the Falcon-H1
    granularity sweep, where \curvefpseven crosses INT8 at G512.}
    \label{fig:inference-quality}
\end{figure*}

At G64, \curvefpseven beats tensorwise FP8 on all four models and remains within
1.32\% perplexity of native. Against rowwise FP8 it wins on Llama and Qwen3.5
and trails by less than 0.016 PPL on Qwen3 and Falcon-H1. Figure~\ref{fig:inference-quality}b
shows the broader scaling trend: INT8 leads narrowly at G128/G256,
\curvefpseven crosses at G512, and its advantage grows under rowwise and
tensorwise sharing. The seven-bit format therefore retains FP8-class quality
while reducing element width and preserving closed-product arithmetic.

\subsection{Rational radix validates the quality--phase control}
\label{sec:experiments:radix}

All principal formats use binary radix. To test the rational generalization, we
vary only the E4C3 radix while holding element width, G512 blocks, and scale
count fixed. A near-binary point adds one predicted phase and improves all four
models slightly. A denser point reaches the best perplexity but needs 32 phases
for only 0.0055--0.0259 lower PPL than binary radix. This confirms the
quality--phase law and motivates the eight-phase operating point; full results
are shown in Table~\ref{tab:radix-frontier}.

\begin{table}[h]
    \caption{E4C3 WikiText-2 perplexity at G512 as radix trades phase count for
    local resolution. Lower is better.}
    \label{tab:radix-frontier}
    \centering
    \small
    \begin{tabular}{lccc}
        \toprule
        & $r=2$ & $r=2^{8/9}$ & $r=2^{3/4}$ \\
        Model & 8 phases & 9 phases & 32 phases \\
        \midrule
        Llama-3-8B   & 6.2160 & 6.2052 & \textbf{6.1901} \\
        Qwen3-8B     & 7.0536 & 7.0426 & \textbf{7.0420} \\
        Qwen3.5-9B   & 7.0159 & 7.0070 & \textbf{7.0034} \\
        Falcon-H1-7B & 5.6654 & 5.6652 & \textbf{5.6599} \\
        \bottomrule
    \end{tabular}
\end{table}

\subsection{CurveFP8 reduces error in all three training GEMMs}
\label{sec:experiments:gemms}

Before pretraining, we capture real forward $XW^{\mathsf T}$,
activation-gradient $\nabla YW$, and weight-gradient $\nabla Y^{\mathsf T}X$
operands from Pythia-410M~\citep{biderman2023pythia} and
Llama-3.2-3B~\citep{grattafiori2024llama3}. Each window is a disjoint 64-token
segment and one real backward pass; three layers per linear role are spaced
through model depth. Under identical tensors, \curvefpeight lowers NMSE in all
36 paired comparisons and by 8.2--11.1\% in aggregate
(Table~\ref{tab:training-gemms}). The gain therefore covers the complete
linear-layer training lifecycle rather than only the forward product.

\begin{table}[h]
    \caption{\curvefpeight/FP8 NMSE on paired training-GEMM windows. Values
    below one favor \curvefpeight; all 36 comparisons are wins.}
    \label{tab:training-gemms}
    \centering
    \small
    \begin{tabular}{lcccc}
        \toprule
        Model & Windows & Forward & $\nabla X$ & $\nabla W$ \\
        \midrule
        Pythia-410M & 8 & 0.8927 & 0.8887 & 0.9012 \\
        Llama-3.2-3B & 4 & 0.9044 & 0.8952 & 0.9180 \\
        \bottomrule
    \end{tabular}
\end{table}

\subsection{CurveFP8 completes three matched 3B-token runs}
\label{sec:experiments:pretraining}

The decisive test trains three matched BF16/FP8/\curvefpeight triplets: nine
128.3M-parameter Llama-style decoders, each on 2,999,943,168 FineWeb-Edu
tokens~\citep{penedo2024fineweb}. The model has 12 layers, width 768,
intermediate width 2560, 12 attention heads, and four K/V heads. We train for
30,517 updates at sequence length 1024 and global batch size 96 using AdamW,
a 1,000-step warmup, and cosine decay. Modes within a triplet share
initialization, token order, optimizer, and schedule. FP8 uses E4M3/E5M2;
\curvefpeight uses E4C3/E5C2 with power-of-two scales. Parameters and optimizer
states remain FP32, while nonlinear operations use BF16.

We evaluate every checkpoint in two ways. First, each model runs in the
arithmetic used during training: BF16, FP8, or \curvefpeight. This is the
deployment result. Second, we run the same learned weights in BF16 for all three
models. This common-arithmetic comparison reveals which training mode learned
the better checkpoint. Within each row of Table~\ref{tab:pretraining-endpoints},
the gap between these two evaluations is the inference quantization loss.

\begin{figure*}[t]
    \centering
    \includegraphics[width=0.96\textwidth]{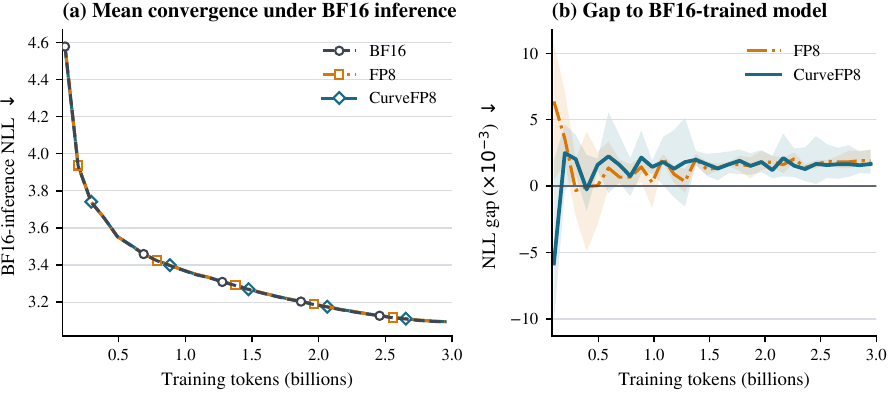}
    \caption{\textbf{\curvefpeight and FP8 track BF16 across three seeds.}
    Left: BF16 evaluation isolates learned checkpoint quality. Right: each
    quantized mode after subtracting seed-matched BF16. Lines are seed means;
    bands span the seed range.}
    \label{fig:pretraining-trajectory}
\end{figure*}

\begin{table}[h]
    \caption{Held-out perplexity after matched 3B-token pretraining. Values are
    three-seed mean $\pm$ sample standard deviation.}
    \label{tab:pretraining-endpoints}
    \centering
    \small
    \begin{tabular}{lcc}
        \toprule
        Training mode & Training-format PPL $\downarrow$ & BF16 PPL $\downarrow$ \\
        \midrule
        BF16 & $22.4899\pm0.0198$ & $22.4899\pm0.0199$ \\
        FP8 & $22.6164\pm0.0130$ & $22.5407\pm0.0182$ \\
        \curvefpeight & $\textbf{22.6054}\pm0.0199$ & $\textbf{22.5366}\pm0.0207$ \\
        \bottomrule
    \end{tabular}
\end{table}

The low-precision endpoints are statistically unresolved. CurveFP8-minus-FP8
perplexity is $-0.0110\pm0.0217$ in training arithmetic and
$-0.0040\pm0.0206$ under BF16 inference, with \curvefpeight winning two of
three seeds in both views. The format-induced penalty is more consistent:
0.0688 versus 0.0758 PPL, lower for \curvefpeight in every seed. The trajectories
also overlap, and all nine lanes complete without divergence or a non-finite
update. These results establish full-run FP8-class training parity, not
statistical superiority.

\subsection{CurveFP8 transfers across seeds and runtime formats}
\label{sec:experiments:downstream}

We next evaluate WikiText-103 and PG-19 OOD language modeling plus zero- and
five-shot suites spanning HellaSwag, PIQA, ARC, WinoGrande, LAMBADA, and BLiMP.
Every cell shares the tokenizer, 1,024-token context, and lm-evaluation-harness
protocol; Appendix~\ref{app:experiments:downstream} gives task and bootstrap
details plus the full training/runtime matrix.

\begin{table}[h]
    \caption{Final checkpoints benchmarks evaluation. Values are three-seed mean
    $\pm$ sample standard deviation; task values are percentages.}
    \label{tab:downstream-native}
    \centering
    \scriptsize
    \begin{tabular}{lccccc}
        \toprule
        Trained & WikiText PPL & PG-19 PPL & Zero-shot & Five-shot & BLiMP \\
        \midrule
        BF16 & $40.384\pm.071$ & $43.728\pm.609$ & $\mathbf{40.128\pm.164}$
             & $\mathbf{43.096\pm.531}$ & $79.786\pm.343$ \\
        FP8 & $40.487\pm.034$ & $43.886\pm.669$ & $39.713\pm.299$
            & $42.446\pm.563$ & $79.597\pm.298$ \\
        \curvefpeight & $\mathbf{40.083\pm.010}$ & $\mathbf{43.132\pm.528}$
            & $39.906\pm.214$ & $42.629\pm.118$ & $\mathbf{79.938\pm.201}$ \\
        \bottomrule
    \end{tabular}
\end{table}

\begin{figure*}[t]
    \centering
    \includegraphics[width=0.95\textwidth]{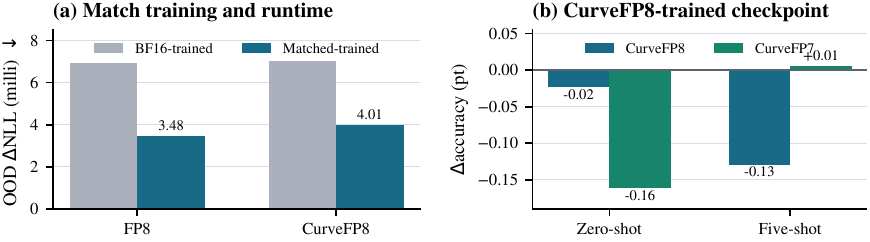}
    \caption{\textbf{Matching training and runtime reduces format-transfer
    loss.} Left: OOD penalty for BF16-trained and matched-trained checkpoints.
    Right: zero- and five-shot changes for \curvefpeight and \curvefpseven on
    the \curvefpeight-trained checkpoint.}
    \label{fig:downstream-transfer}
\end{figure*}
\FloatBarrier
Table~\ref{tab:downstream-native} evaluates learned checkpoint quality under a
common BF16 runtime. \curvefpeight has the best mean perplexity on both OOD
corpora and the best BLiMP mean, while BF16 leads the aggregate zero- and
five-shot suites. Figure~\ref{fig:downstream-transfer} then isolates runtime
format sensitivity: matching training and runtime nearly halves the OOD penalty
for both FP8 and \curvefpeight, and the \curvefpeight-trained checkpoint
transfers cleanly to seven-bit deployment. Across runtime formats,
\curvefpeight improves WikiText-103 in all 12 paired comparisons; PG-19 and task
outcomes are mixed. This supports transfer parity with one consistent OOD
advantage, not broad downstream superiority.

%% file: sections/discussion.tex
\section{Arithmetic Realization and Hardware Scope}
\label{sec:deployment}

The contribution is primarily algorithmic: the element fields compose exactly,
and Equation~\ref{eq:phase-count} turns each radix choice into an explicit
accumulation contract. Sign XOR, curve-index addition, carry, exponent addition,
and binary scale transport determine every nonzero product without lookup or
projection. Products then route to one of $H$ phases and one binary bin, with
counter width fixed by dot length. A banked histogram can merge equal addresses
through signed popcounts; a per-phase Kulisch layout shifts updates directly
into wide accumulators. Both realize the same semantics before fixed-coefficient
phase reduction. Thus the datatype fixes operations, state bounds, and outputs
while leaving lane count, bank organization, and pipeline placement to the
architecture. Appendix~\ref{app:algorithm} gives the full derivation.

\paragraph{Tensor-core-like spatial architecture.}
A tile is a small matrix-multiplication block intended to be replicated inside
an accelerator. To test whether the closed product maps to useful parallel
arithmetic, we build such a tile rather than a scalar dot-product unit. It
computes a $4\times4$ output block
$C_{i,j}=\sum_{k=0}^{K-1}A_{i,k}B_{k,j}$ as a stream of outer products over the
reduction dimension $K$. On each cycle it accepts four 8-bit values
$A_{0:3,k}$ and four 8-bit values $B_{k,0:3}$, broadcasts each $A$ value across
one row and each $B$ value down one column, and forms all 16 products
$A_{i,k}B_{k,j}$. The 16 processing elements each retain one running output
$C_{i,j}$ locally; this output-stationary dataflow avoids reducing the products
to a single scalar. A start signal clears the partial sums for a new output
block, a last signal captures them after $K$ cycles, and the completed matrix is
read out one row at a time. Both tiles can accept one outer product per cycle,
so their common peak throughput is 16 multiply--accumulate operations (MACs)
per cycle.

The comparison holds the surrounding arithmetic constant. Both tiles consume
the same eight input bytes per cycle, use the same matrix controls and row-drain
interface, and accumulate without intermediate rounding into 16 signed 64-bit
fixed-point registers with 32 fractional bits (Q32). This common accumulator
isolates product hardware; it is not intended to reproduce a commercial tensor
core's FP32 accumulator. CurveFP forms a product by adding the two magnitude
ranks, selecting one fixed phase coefficient, applying the sign, and shifting
the result into Q32. The FP8 reference instead decodes E4M3FN fields, inserts
the hidden bit or handles a subnormal, multiplies the significands, applies the
sign, and aligns the result into the same Q32 accumulator. Thus the experiment
isolates the datatype-specific product path; memories, tensor scales, and final
output conversion are excluded from both designs.

We independently test two pipeline choices for each datatype. P0 connects
product formation directly to the accumulator; P1 inserts one register between
them. That register shortens the combinational path and adds one cycle of result
latency, but both variants still accept a new outer product every cycle. We
place and route all four candidates under the same 2.0~ns clock and I/O
constraints, then select the shallowest variant that closes timing. Setup slack
is required arrival time minus actual arrival time after routing: positive slack
means data arrives before the clock deadline, whereas negative slack is a timing
violation. This criterion compares equal semantics and throughput without
forcing CurveFP to carry a register that its product path may not need.

We placed and routed these tiles in an open Nangate45 flow
(Table~\ref{tab:spatial-hardware}). At 500 MHz,
\curvefpeight closes without a product register, whereas FP8 requires one; the
resulting \curvefpeight tile is 4.6\% smaller at equal peak throughput. Adding
the register produces the P1 operating point. When both designs use P1,
\curvefpeight remains 4.1\% smaller and has 0.426~ns setup slack versus
0.302~ns for FP8. This extra positive margin suggests headroom for a tighter
timing target, hint for higher clock speed. These measurements are a feasibility check rather than a complete
accelerator comparison: memories, interconnect, scale handling, boundary
conversion, and workload-dependent utilization remain outside both tiles. The
complete pipeline sweep, RTL verification, open tool flow, and limitations are
in Appendix~\ref{app:hardware}.
\begin{table}[h]
	\caption{Preliminary $4\times4$ spatial-tile results at 500 MHz. P0/P1
	denotes zero/one product register; the appendix gives the full sweep.}
	\label{tab:spatial-hardware}
	\centering
	\small
	\begin{tabular}{lrrrr}
		\toprule
		Tile & Product reg. & Area ($\mu\mathrm{m}^2$) & Setup slack (ns) & pJ/MAC \\
		\midrule
		\curvefpeight E4C3 P0 & \textbf{0} & \textbf{49,282.9} & $+0.064$ & 4.78 \\
		\curvefpeight E4C3 P1 & 1 & 49,540.1 & $\textbf{+0.426}$ & 4.78 \\
        FP8 E4M3FN P0         & 0 & 53,245.5 & $-0.045$ & -- \\
		FP8 E4M3FN P1         & 1 & 51,646.8 & $+0.302$ & 4.75 \\
		\bottomrule
	\end{tabular}
\end{table}

%% file: sections/conclusion.tex
\section{Conclusion}
\label{sec:conclusion}

\curvefp shows that low-precision representation and product algebra can be
designed together. Its interleaved logarithmic curves give exact index-domain
products, and the phase law makes accumulation cost explicit. \curvefpseven
reaches the FP8 quality regime at seven bits; \curvefpeight improves every
paired training-GEMM diagnostic and matches FP8 across three 3B-token training
triplets. The learned checkpoints retain downstream capability across runtime
formats. A preliminary routed tile adds a hardware hint: at 500 MHz,
\curvefpeight is 4.6\% smaller and one product stage shallower than
timing-closing FP8, or 4.1\% smaller with the stage restored for additional
timing slack. Memory, interconnect, scale transport, and workload effects remain
open. Thus the present evidence supports a coherent FP8-class training path, a
7-bit inference point, and an area-efficient closed-product
realization.

%% file: sections/appendix.tex
\section{Algorithmic Details}
\label{app:algorithm}

\subsection{Closed-product addressing and accumulator bounds}
\label{app:algorithm:accumulation}

Let $g=\gcd(p,qK)$, $n_x=e_xK+k_x$, $n_w=e_wK+k_w$, and
$u=p(n_x+n_w)$. Each nonzero product contributes
\begin{equation}
    \delta=(-1)^{s_x\oplus s_w},\qquad
    h=\frac{u\bmod qK}{g},\qquad
    t=a_x+a_w+\left\lfloor\frac{u}{qK}\right\rfloor
\end{equation}
to integer bin $A_{h,t}$; the zero sentinel suppresses the update. Multiples of
$g$ form the additive subgroup generated by $p$ in $\mathbb{Z}_{qK}$, whose
order is $H=qK/g$. This proves that all products fall into the $H$ phases used
in Equation~\ref{eq:binned-accumulation}.

For binary radix, $p=q=1$, so
\begin{equation}
    h=(k_x+k_w)\bmod K,\qquad
    t=a_x+a_w+e_x+e_w+\left\lfloor\frac{k_x+k_w}{K}\right\rfloor.
\end{equation}
The lane logic is therefore zero detection, sign XOR, one $(C+1)$-bit curve
addition, and signed exponent addition. A length-$D$ dot requires
$1+\lceil\log_2(D+1)\rceil$-bit signed counters. Equal addresses can be merged
with signed popcounts in a histogram implementation, or shifted directly into
one wide accumulator per phase. Both layouts produce the same phase sums before
the fixed-coefficient reduction and final requantization.

\subsection{Shared scales and training operators}
\label{app:algorithm:operators}

A block of $G$ elements with an $S$-bit power-of-two scale has effective rate
\begin{equation}
    R=1+E+C+\frac{S}{G}\quad\text{bits per value}.
\end{equation}
Static operands select a representable scale by reconstruction-MSE search;
dynamic training operands use ceil-absmax scaling. Because the scale is binary,
it changes only the exponent-bin address and introduces no product phase.

For each non-head linear layer, the forward pass evaluates
$Q_A(\bm{X})Q_W(\bm{W})^{\mathsf T}$. The backward pass reuses the quantized
forward operands and quantizes the output gradient, giving
$Q_G(\nabla\bm{Y})Q_W(\bm{W})$ for $\nabla\bm{X}$ and
$Q_G(\nabla\bm{Y})^{\mathsf T}Q_A(\bm{X})$ for $\nabla\bm{W}$. Parameters,
optimizer states, normalization, nonlinear operations, and the tied output head
remain at higher precision.

\section{Experimental Details}
\label{app:experiments}

\subsection{Inference and GEMM diagnostics}

Post-training quantization uses the complete WikiText-2 test split
\citep{merity2016wikitext} and a 2,048-token rolling-loglikelihood protocol.
Within each model, all methods share
weights, output-head exclusion, and K/V precision. Llama-3 and Qwen3 keep dense
K/V at common tensorwise FP8 so the comparison isolates weight/activation
arithmetic.

The training-GEMM diagnostic records forward $XW^{\mathsf T}$,
activation-gradient $\nabla YW$, and weight-gradient $\nabla Y^{\mathsf T}X$
operands from Pythia-410M and Llama-3.2-3B. Each window is a disjoint 64-token
WikiText-2 segment with one real backward pass. Three layers per linear role are
spaced through model depth, yielding 12 sampled Pythia linears and 21 sampled
Llama linears per window. CurveFP8 lowers aggregate NMSE by 8.2--11.1\% and
wins all 36 paired model/window/GEMM comparisons reported in
Table~\ref{tab:training-gemms}.

\subsection{Pretraining protocol}

The matched experiment trains three independent BF16/FP8/CurveFP8 triplets.
Each lane is a 12-layer, 128.3M-parameter Llama-style decoder with width 768,
intermediate width 2560, 12 attention heads, and four key/value heads. Runs use
2,999,943,168 FineWeb-Edu tokens~\citep{penedo2024fineweb}, sequence length
1024, global batch size 96, and 30,517 updates. AdamW uses a 1,000-step warmup
followed by cosine decay from
$6\times10^{-4}$ to $6\times10^{-5}$. Modes within a triplet share
initialization, token order, optimizer, and schedule. Parameters and optimizer
states remain FP32; nonlinear operations use BF16. Training disables K/V
caching, and each lane applies its forward format to K/V projection linears.

\subsection{OOD and downstream evaluation}
\label{app:experiments:downstream}

\begin{figure*}[t]
    \centering
    \includegraphics[width=0.95\textwidth]{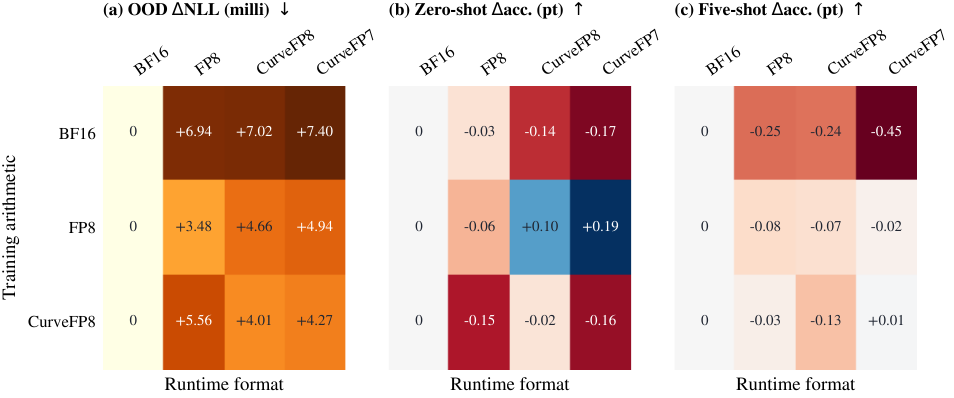}
    \caption{Complete three-seed format-transfer matrix. Each row fixes the
    training arithmetic and each column the runtime format; values are relative
    to BF16 runtime on the same checkpoint.}
    \label{fig:downstream-transfer-full}
\end{figure*}

OOD language modeling uses the complete WikiText-103 and PG-19 test
sets~\citep{merity2016wikitext,rae2019compressive}. The zero-shot suite contains
HellaSwag, PIQA, ARC-Easy/Challenge, WinoGrande, LAMBADA, and
BLiMP~\citep{zellers2019hellaswag,bisk2020piqa,clark2018arc,
sakaguchi2019winogrande,paperno2016lambada,warstadt2020blimp}; the five-shot
suite repeats all but ARC-Challenge and BLiMP. Evaluation uses
lm-evaluation-harness 0.4.11, the training tokenizer,
1,024-token context, and 10,000 bootstrap iterations.
Figure~\ref{fig:downstream-transfer-full} gives all training/runtime-format
deltas; Table~\ref{tab:downstream-native} in the main paper gives absolute
BF16-runtime checkpoint quality.

CurveFP8 gives the strongest paired result on WikiText-103, lowering perplexity
in all 12 seed/runtime-format comparisons. PG-19 favors CurveFP8 in 8/12
comparisons but has larger seed variation; zero- and five-shot means favor it
in 7/12 and 6/12 comparisons and remain close across runtime formats. The
largest conservative single-seed combined-error score is $|z|=1.43$. These
results support one consistent OOD advantage and broad transfer parity, not
universal downstream superiority.

\section{Spatial-Tile Hardware Microbenchmark}
\label{app:hardware}

\subsection{Arithmetic and tile contract}

The hardware check compares CurveFP8 E4C3 with FP8 E4M3FN in matched
$4\times4$ output-stationary tiles. Four A bytes and four B bytes are broadcast
per cycle, forming 16 products that enter 16 independent signed 64-bit Q32
accumulators. Start/last controls delimit variable-$K$ outer-product streams;
completed outputs drain as four 256-bit rows. Both tiles therefore accept 16
MACs per cycle. SRAM, NoC, scale generation and storage, scheduling, larger
tiling, result buffering, and BF16/FP16 boundary conversion are excluded from
both designs.

CurveFP packs sign in bit 7 and magnitude rank $m$ in bits 6:0, with $m=0$
reserved for zero. For nonzero operands, $q=m-64$ and the product coordinate is
$q_p=m_a+m_b-128$. E4C3 splits $q_p=8e+k$; the implementation selects one Q16
coefficient $\operatorname{round}(2^{k/8}2^{16})$, shifts it into Q32, applies
the product sign, and accumulates. FP8 performs E4M3FN field decoding,
subnormal handling, significand multiplication, sign application, and binary
alignment into the same Q32 accumulator. Neither design rounds intermediate
products or partial sums.

P0 places no register between product formation and accumulation; P1 adds one
product/control register. The comparison independently selects the shallowest
variant that closes the common 2.0 ns constraints rather than forcing equal
latency. Table~\ref{tab:spatial-hardware-full} records all four routed
candidates.

\begin{table}[h]
    \caption{Complete Nangate45 pipeline sweep. WS and TNS are in ns; ``--''
    denotes an unmeasured operating point.}
    \label{tab:spatial-hardware-full}
    \centering
    \scriptsize
    \begin{tabular}{lrrrrrrr}
        \toprule
        Tile & Reg. & Area ($\mu\mathrm{m}^2$) & Cells & Setup WS & Setup TNS & Hold WS & pJ/MAC \\
        \midrule
        CurveFP8 P0 & 0 & 49,282.9 & 37,999 & $+0.064$ & 0 & $+0.104$ & 4.78 \\
        CurveFP8 P1 & 1 & 49,540.1 & 33,868 & $+0.426$ & 0 & $+0.104$ & 4.78 \\
        FP8 P0      & 0 & 53,245.5 & 41,077 & $-0.045$ & $-1.223$ & $+0.105$ & -- \\
        FP8 P1      & 1 & 51,646.8 & 35,716 & $+0.302$ & 0 & $+0.106$ & 4.75 \\
        \bottomrule
    \end{tabular}
\end{table}

\subsection{Verification, tools, and power method}

All 256 E4M3FN byte patterns were checked against PyTorch 2.6, and all 16,384
CurveFP E4 magnitude pairs were checked exhaustively. Selected RTL and final
routed netlists pass 64 seeded variable-$K$ matrices spanning 1,010 compute
cycles and 1,024 checked outputs per datatype. Both selected routes have zero
setup/hold TNS, final detailed-route DRC errors, antenna violations, and final
flow errors.

The reproducible flow uses OSS CAD Suite 2026-08-11 (Yosys 0.68, Verilator
5.051, Icarus 14, and Slang 11), OpenROAD Flow Scripts revision
\texttt{56496f3}, OpenROAD revision \texttt{ab6fd26}, and the public Nangate45
library.

Power uses a matched 4,096-cycle, $K=64$ stream at 500 MHz with identical
finite input bytes, matrix controls, and row-drain timing. OpenSTA reads each
final ODB, SDC, extracted SPEF, and gate-level VCD; all routed pins are
annotated. The functional cell simulation is zero-delay and omits
timing-dependent glitches, and only one physical seed is measured.
Consequently, the paper claims area and product-stage flexibility only;
frequency scaling and application throughput require system-level study.